\documentclass[%
preprint,
 amsmath,amssymb,
 aps,
 pre,
floatfix,
]{revtex4-1}

\usepackage{graphicx}
\usepackage{dcolumn}
\usepackage{bm}
\usepackage{wasysym} 
\usepackage[cdot,squaren]{SIunits} 

\usepackage{nicefrac}

\newcommand{\caap}[1]{\operatorname{CAAP}\left(#1\right)}
\newcommand{\gait}[1]{\operatorname{\mathcal{#1}}}
\newcommand{\TO}{{\tiny\mars}}

\begin{document}

\preprint{APS/123-QED}

\title{Exploiting the Passive Dynamics of a Compliant Leg to Develop Gait Transitions}

\author{Harold Roberto Martinez Salazar }
	\homepage{http://ailab.ifi.uzh.ch/martinez/}
	\email{martinez@ifi.uzh.ch}

\author{Juan Pablo Carbajal}%
	\homepage{http://ailab.ifi.uzh.ch/carbajal/}
 \email{carbajal@ifi.uzh.ch, both authors can be contacted regarding the content of the paper}
\affiliation{%
 Artificial Intelligence Laboratory, Department of Informatics,\\
University of Zurich\\
Andreasstrasse 15 8050 Zurich Switzerland
}%




\date{\today}

\begin{abstract}
 
\begin{description}
\item[Abstract]
In the area of bipedal locomotion, the spring loaded inverted pendulum (SLIP) model has been proposed as a unified framework to explain the dynamics of a wide variety of gaits. In this paper, we present a novel analysis of the mathematical model and its dynamical properties.  We use the perspective of hybrid dynamical systems to study the dynamics and define concepts such as partial stability and viability. With this approach, on the one hand, we identified stable and unstable regions of locomotion.  On the other hand, we found ways to exploit the unstable regions of locomotion to induce gait transitions at a constant energy regime. Additionally, we show that simple non-constant angle of attack control policies can render the system almost always stable.
 
\end{description}
\end{abstract}

\pacs{Valid PACS appear here}
\maketitle


\section{\label{sec:level1}Introduction}

\begin{figure}[b]

\includegraphics{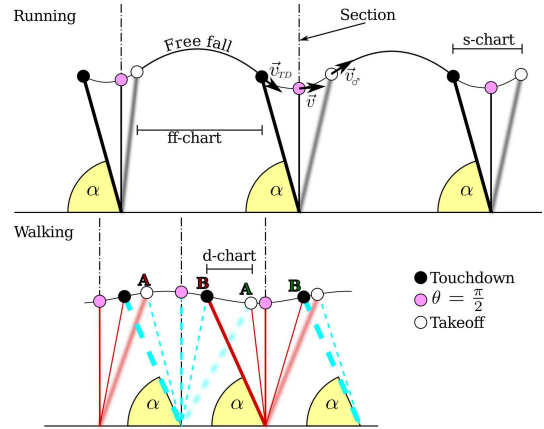}
\caption{\label{fig:ModelDiagram} (Color online) Illustration of the evolution of the SLIP model for running and walking. The mass is represented with a filled circle. The color of the fill indicates touchdown event (black), takeoff event (white), and the crossing of the section (pink (grey)). The landing leg is pictured with a thick solid line, and the leg at takeoff is represented with a blurred line. Due to the passive properties of these models, control is necessary only during the swing of the leg, i.e. during free fall while running and from point A to B while walking.}
\end{figure}

One of the most accepted mathematical models for bipedal running is the spring loaded inverted pendulum (SLIP, for an extensive review see\cite{HolmesSIAM.06}). In a similar fashion, the rigid inverted pendulum has been extensively used to model bipedal walking\cite{Mochon1980}. In 2006, Geyer et al.\cite{Geyer2006} propose the SLIP model as a unifying framework to describe walking as well as running. The unified perspective proves useful for accurately explaining data from human locomotion\cite{Geyer2006}. Additionally, it allows describing both gaits (walking and running) in terms of dynamical entities observed in a discrete map, obtained by intersecting the trajectories of the system with a predefined section of lower dimension. Geyer associates these entities with limit cycles of the {\it hybrid dynamical system}\cite{Guckenheimer95,Cortes2008} and named their attracting behavior as {\it self-stabilization}. Though the nature of the observed dynamical properties is not yet clarified, those results emphasize that bipedal locomotion may be dictated solely by the mechanics of the system. As a consequence, the control necessary for locomotion is thus reduced to the swing phase of the leg, showed in  Fig.~\ref{fig:ModelDiagram} between points A and B. The most popular control policy is to produce touchdowns at constant angle of attack $\alpha$ ($\caap{\alpha}$), i.e. the angle spanned by the landing leg and the horizontal.

In the last decade, many energy-efficient bipedal walking machines have been developed. Through careful design, they exploit the passive dynamics of their own body to move forward, requiring little control or none\cite{McGeer1990, Collins2001,MartijnWisseandJanvanFrankenhuyzen2006, Collins2005,Geng2006}. However, the construction of bipedal machines capable of exploiting passive dynamics in different gaits remains an unsolved engineering challenge. In this context, Geyer et al.\cite{Geyer2006} report that, in the SLIP model, it is not possible to have multiple gaits at the same energy. The results are based on simulations that do not cover all possible initial conditions of the system. In addition, Rummel et al.\cite{Rummel2010} prove that walking and running is possible at the same energy level. They use a new map that allows comparing different gaits with ease. The map is defined at the vertical plane crossing the landing point of the foot (Fig.~\ref{fig:ModelDiagram}). In this way, they find the self-stable regions, but their intersection is empty. To concretize these ideas, let us describe this region for the running map $\gait{R}$.

\begin{equation}
E_\infty^R = \lbrace x \vert\; x \in \mathcal{S} \wedge \left( \exists \alpha \vert\; x = \gait{R}_\alpha\left(x\right)\right)\rbrace,
\label{eq:StableRegion}
\end{equation}

\noindent where the subscript in $\gait{R}_\alpha$ denotes running using $\caap{\alpha}$ and $\mathcal{S}$ denotes the section where the map is defined. Therefore, if for different gaits these stable regions do not intersect, e.g. $E_\infty^R \cap E_\infty^W = \emptyset$,  we conclude that a transition between the two gaits cannot occur if the system is to remain in these regions. In other words,

\begin{equation}
\begin{split}
x \in E_\infty^R \: \wedge \: & y \in E_\infty^W \quad \Rightarrow \\
& \gait{R}_\alpha \left(y\right) \notin E_\infty^R \: \wedge \: \gait{W}_\beta \left(x\right) \notin E_\infty^W \quad \forall \: \alpha,\,\beta.
\end{split}
\end{equation}

In this study, we will show how transitions between gaits are found at points outside these stable regions. The transitions require the selection of the angle of attack; therefore CAAP's are not suitable for this task. We will also show evidence indicating that it is possible to find an angle of attack $\theta$ that maps a point into a stable region, e.g. $x \notin E_\infty^R \: \wedge \: \left(\exists\:\theta, y \;\vert \: y \neq x,\; y \in E_\infty^R, \; y = \gait{R}_{\theta}(x)  \right)$. Additionally, we introduce the concepts of {\it partial stability} and {\it viability} that will be useful in the construction of the transitions presented herein.

This paper is organized as follows. In section \ref{sec:MatandMeth}, we describe the models used for our simulations, their representation in state variables and the definition of the discrete map. Next, in section \ref{sec:Results}, we introduce the new concepts, and we show the regions where the transitions between gaits exist. Later, in section \ref{sec:discussion}, we discuss about the requirements of a controller for the system and the implications for robot design and bipedal locomotion. We conclude the paper in section \ref{sec:conclusion} with our conclusion.

\section{Methods\label{sec:MatandMeth}}

As explained previously, we use the SLIP model to study bipedal gaits. We adopt the framework in \cite{JuergenRummel2009}, which is described in the language of hybrid dynamical systems. Therefore, we reintroduce some notation and definitions.

To represent the different phases of a gait, the model is segmented into three sub-models. We will call these sub-models {\it charts}\cite{Guckenheimer95} or phases see Fig.\ref{fig:ModelDiagram}. Each chart represents the motion of a point mass under the influence of: only gravity (ff-chart or flight phase), gravity and a linear spring (s-chart or single stance phase),  gravity and  two linear springs (d-chart or double stance phase). The point mass represents the body of the agent and the massless linear springs model the forces from the legs (Fig.\ref{fig:ModelDiagram}). A trajectory switches from one chart to another when some real valued functions evaluated on it cross zero ({\it event functions}\cite{Guckenheimer95,Piiroinen08}). We define a running gait as a trajectory that switches from the s-chart to the ff-chart and back to the s-chart. A walking gait is defined as a trajectory that switches from the s-chart to the d-chart and back again to the s-chart. Switches from the ff-chart to d-chart or vice versa are not included in this study.

\subsection{Equations of motion in each chart}
The motion in all the charts is governed by a system of ordinary differential equations:

\begin{equation}
\dot{\vec{X}}  = \vec{F_i}\left(\vec{X}\right),
\end{equation}

\noindent where $\vec{X}$ is the vector of state variables and $\vec{F_i}$ is a force function characteristic of each chart. Since all forces are conservative, the energy of the system is constant. For the ff-chart the state is described by the Cartesian coordinates of the position of the point mass and its velocity $\vec{X}_{ff} = \left(x,y,v_x,v_y\right)^T$,
\begin{equation}
\dot{\vec{X}}_{ff}  = \begin{pmatrix}v_x\\ v_y\\0\\-g\end{pmatrix},
\label{eq:ff-chart}
\end{equation}

\noindent where $g$ is the acceleration due to gravity.

The state in the s-chart is represented in polar coordinates $\vec{X_s} = \left(r,\theta,\dot{r},\dot{\theta}\right)^T$, where $r$ is the length of the spring and $\theta$ is the angle spanned by the leg and the horizontal, growing in clockwise direction. Thus, the equations of motion are:
\begin{equation}
\dot{\vec{X}}_{s}  = \begin{pmatrix}\dot{r}\\ \dot{\theta}\\ \frac{k}{m}\left( r_0-r\right)+ r\dot{\theta}^2-g\sin\theta\\ -\frac{1}{r}\left(2\dot{r}\dot{\theta} +g\cos\theta\right) \end{pmatrix}.
\label{eq:s-chart}
\end{equation} 
\noindent It is important to note that $\theta(t_{TD}) = \alpha$, i.e. the angular state at the time of touchdown is equal to the angle of attack. The parameter $r_0$ defines the natural length of the spring.

In the d-chart the state is also represented in polar coordinates $\vec{X_d} = \left(r,\theta,\dot{r},\dot{\theta}\right)^T$, with the origin of coordinates in the new touchdown point. The motion is described by:
\begin{eqnarray}
&&\dot{\vec{X}}_{d}  = \begin{pmatrix}
\dot{r}\\ 
\dot{\theta}\\ 
\begin{split}
\frac{k}{m}\left[(r_0-r) + \left(1-\frac{r_0}{r_{\TO}}\right)(x_{\TO}\cos\theta - r) \right] \\ +r\dot{\theta}^2-g\sin\theta
\end{split}
\\ 
\begin{split}
-\frac{1}{r}\left[\frac{k}{m}\left(1-\frac{r_0}{r_{\TO}}\right)x_{\TO}\sin\theta + 2\dot{r}\dot{\theta}+g\cos\theta \right] 
\end{split}
\end{pmatrix}\\
\label{eq:d-chart}
&& r_{\TO} = \sqrt{r^2 + x_{\TO}^2-2rx_{\TO}\cos\theta}, \label{eq:rTO}
\end{eqnarray} 
\noindent where $x_{\TO}$ is the horizontal distance between the two contact points and $r_{\TO}$ is the length of the back leg.

\subsection{Event functions}
Event functions are functions on the phase space of the system. An event occurs when the trajectory of the system intersects a level curve of the event function. At the time of the event, the current state of the system is mapped to the state of another chart. Some event functions are parameterized with the angle of attack and the natural length of the springs. 

Switches from the ff-chart to the s-chart are defined by:
\begin{equation}
\mathcal{F}_{ff \rightarrow s}\left(\vec{X}_{ff},\alpha,r_0\right) : \begin{cases}  y - r_0\cos\alpha = 0\\ v_y < 0 \end{cases},
\label{eq:ff2s}
\end{equation}
\noindent which means that the mass is falling and the leg can be placed at its natural length with angle of attack $\alpha$. Therefore, the motion is now defined in the s-chart. The switch in the other directions is simply:
\begin{equation}
\mathcal{F}_{s \rightarrow ff}\left(\vec{X}_{s},r_0\right) :  r - r_0 = 0.
\end{equation}
\noindent These are the only two event functions involved in the running gait. The map from one chart to the other is defined by:
\begin{equation}
\quad x = -r\cos \theta \qquad  y = r\sin\theta.
\end{equation}
\noindent It is important to have in mind that the origin of the s-chart is always at the touchdown point.

For the walking gait, we have to consider switches between single and double stance phases. From the s-chart to the d-chart, we have:
\begin{equation}
\mathcal{F}_{s \rightarrow d}\left(\vec{X}_{s}, \alpha, r_0\right) : \begin{cases}  r\sin\theta - r_0\cos\alpha = 0\\ \theta > \frac{\pi}{2} \end{cases},
\end{equation}
\noindent which is similar to (\ref{eq:ff2s}) with the additional condition that the mass is tilted forward. Additionally, if we consider the sign of the radial speed, we differentiate between walking gait $\gait{W}$ with $\dot{r}<0$ and Grounded Running gait $\gait{GR}$, with $\dot{r}>0$.

The switch from the double stance phase to the single stance phase is defined by:

\begin{equation}
\mathcal{F}_{d \rightarrow s}\left(\vec{X}_{d}, r_0\right) : r_{\TO} - r_0 = 0,
\end{equation}
\noindent with $r_{\TO}$ as defined in (\ref{eq:rTO}). The map from the d-chart to the s-chart is the identity. In the other direction we have:
\begin{eqnarray}
&&r_d = r_0 \quad  \theta_d = \alpha, \\
&&x_{\TO} = r_0 \cos\alpha - r_s\cos\theta_s ,
\end{eqnarray}
\noindent where the subscripts indicate the corresponding chart.

If the system falls to the ground ($y \leq 0$), attempts a forbidden transition (e.g. d-chart to ff-chart), or renders $v_x <0$ (motion to the left,``backwards''), we consider that the system fails.

\subsection{Simulation of the dynamics\label{sec:simulation}}
The state of the model is observed when the trajectory of the system intersects the section defined by $\mathcal{S} :\theta = \nicefrac{\pi}{2}$. In this way, the map $\gait{R}_\alpha : \mathcal{S} \rightarrow \mathcal{S}$ transforms points through the evolution of the system from the s-chart to the ff-chart and back again to the s-chart using an angle of attack $\alpha$. Similarly, the map $\gait{W}_\alpha : \mathcal{S} \rightarrow \mathcal{S}$ transforms points through the evolution of the system from the s-chart to the d-chart and back again to the s-chart using an angle of attack $\alpha$. 

All initial conditions are given in the $\mathcal{S}$ section and in the s-chart, i.e. only one leg touching the ground and oriented vertically. Moreover, all the initial conditions are given at the same total energy. The results are visualized using the values of the length of the spring $r$ and the radial component of the velocity which, in $\mathcal{S}$, equals the vertical speed $\dot{r} = v_y$ ($v_x$ is obtained from these values and the equation of constant energy). It is important to note that all possible values of $r$, $v_y$ and $v_x$, for a given value of the total energy $E$, lay on an ellipsoid. Besides, there is a transformation that maps the ellipsoid to a sphere. This can be shown as follows: the total energy in the section is,
\begin{equation}
\label{eq:energy}
E = \frac{1}{2}k\left(r_0-r\right)^2 + \frac{1}{2}m\left(v_x^2 + v_y^2\right) + mgr
\end{equation}

Defining the parameters
\begin{eqnarray}
L &=& \sqrt{\frac{2}{k}\left[E - mg\left(r_0 -\frac{mg}{2k}\right)\right]},\\
\omega &=& \sqrt{\frac{k}{m}},
\end{eqnarray}
\noindent the new variables 
\begin{eqnarray}
\hat{v}_x &=&  \frac{v_x}{\omega},\\
\hat{v}_y &=&  \frac{v_y}{\omega},\\
\hat{r} &=& r- \left(r_0 - \frac{mg}{k}\right),
\end{eqnarray}
transform equation (\ref{eq:energy}) into,
\begin{equation}
\label{eq:sphere}
L^2 = \hat{v}_x^2 + \hat{v}_y^2 + \hat{r}^2
\end{equation}
\noindent which defines a sphere. Therefore, all initial conditions of $\hat{r}$ and $\hat{v}_y$ with constant energy, are defined inside a circle. A Delaunay triangular mesh was created in the circle with $65896$ initial conditions as vertices ($131245$ triangles). Each vertex was transformed using $\gait{R}_\alpha$, $\gait{GR}_\alpha$ and $\gait{W}_\alpha$ with $400$ values of $\alpha \in [55\degree,90\degree]$. To compute the evolution of an arbitrary initial condition, we used bilinear interpolation in the triangles of the mesh.

The model implementation and data analysis were carried out in MATLAB(2009, The MathWorks), GNU Octave\cite{octave2002} and Matplotlib\cite{matplotlib}. Simulations were run for constant energy, using the step variable integrator ode45 (relative tolerance: $1\times 10^{-6}$ and absolute tolerance: $1\times 10^{-8}$). Table \ref{tab:params} shows the values of the parameters used.

\begin{table}

\caption{\label{tab:params} Values used for the simulations presented in this paper.} 
\begin{ruledtabular}
\begin{tabular}{|l|c|c|}
Description & Name & Value \\
Mass & $m$ & \unit{80}\kilogram \\
Elastic constant of linear springs & $k$ & \unit{15} \kilo\newton\metre \\
Rest length of linear springs & $r_0$ & \unit{1}\metre \\
Total energy & $E$ & \unit{820}\joule \\
Acceleration due to gravity & $g$ & \unit{9.81} \metre\per\square\second \\
Angle of Attack & $\alpha$ & from \unit{55}\degree to \unit{90}\degree \\
\end{tabular}
\end{ruledtabular}

\end{table}

\section{Results\label{sec:Results}}
In this section, we present the results of the analysis on the data collected from the models as described in section \ref{sec:simulation}. Aiming to define a controller, we introduce some important properties of the dynamics of each gait, namely finite stability for a given CAAP and viability. 

\subsection{Finite stability and Viability\label{sec:stabilityviability}}
Finite stability describes the set of initial conditions where the system can do a maximum amount of steps (sequential applications of the map) before failing, using CAAP. For example, we can define for $\gait{W}$

\begin{equation}
E_n^W = \lbrace x \vert\; x \in \mathcal{S} \wedge \left( \exists \alpha \vert\; y = \gait{W}_\alpha^n \left(x\right), \; n \geq 1, \; y \in \mathcal{S} \right)\rbrace.
\end{equation}

\noindent That is, at a given state $x=(r,v_y)$ in $\mathcal{S}$ there is a $\caap{\alpha}$ such that the system can do at most $n$ steps before failing. The region $E_0^W$ are all the points in the section where applying $\gait{W}$ produces a failure. The existence of $E_n^W$ implies that a controller of the system may not need to take a decision at each step. In addition, the controller may exploit this alleviation by planning future angles of attack. Viability describes how easy is to choose the future angle of attack. The level of ease is measured in terms of the size of the interval of angles that can be chosen to avoid a failure of the system. For the running gait this region is defined as:

\begin{equation}
\begin{split}
V^R\left(\Delta\alpha\right) = & \lbrace x \vert \; x \in \mathcal{S} \wedge \\ & \left( \exists \alpha \in I_\alpha, \; \Vert I_\alpha \Vert \geq \Delta\alpha \; \vert\; y = \gait{R}_\alpha \left(x\right), \; y \in \mathcal{S} \right)\rbrace,
\end{split}
\end{equation}

\noindent where $I_\alpha$ denotes a real interval and $\Vert\cdot \Vert$ measures its length. In a real system, it is required that a viable angle of attack exists for a definite interval, since real sensors and actuators have a finite resolution and are affected by noise.

\begin{figure*}[htb]
\includegraphics[width=0.8\textwidth]{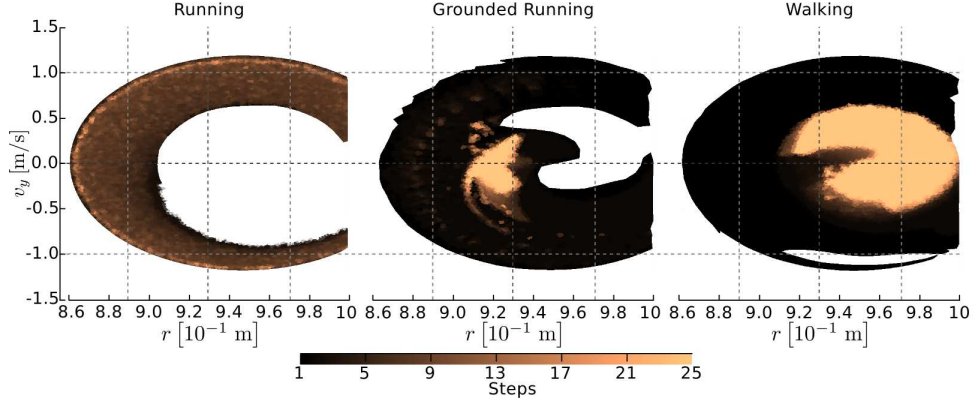}
\caption{\label{fig:NMSRegion} (Color online) Finite stability regions. The figures show initial conditions for $\gait{R}$, $\gait{GR}$ and $\gait{W}$ that can do multiple steps under CAAP before failing. A region in white corresponds to $E_0^i$ for gait $i$.}
\end{figure*}

Fig.~\ref{fig:NMSRegion} shows the finite stability regions for each gait. The stable region of $\gait{R}$, as reported in \cite{JuergenRummel2009} ($v_y = 0$) is not visible. Although $E_\infty^R$ may have some area of attraction, due to the resolution we used for the angles of attack (described in section \ref{sec:simulation}) we do not see it in our results. Based on results not presented here, we estimate that the resolution in the angle of attack to detect such basin for the current energy is $\sim 10^{-4}$. In despite of the low resolution in the angles, the system can perform an average of $10$ steps in $\gait{R}$, and at least $25$ steps (maximum calculated) in $\gait{GR}$ and $\gait{W}$. This means that running is more difficult at this energy level than the other two gaits. Particularly for $\gait{GR}$ and $\gait{W}$, we see that there is a plateau with the maximum number of steps. This is the evidence of the self-stable regions of these gaits, and the plateau is related to the basing of attraction of that region.

\begin{figure*}[htb]
\includegraphics[width=0.8\textwidth]{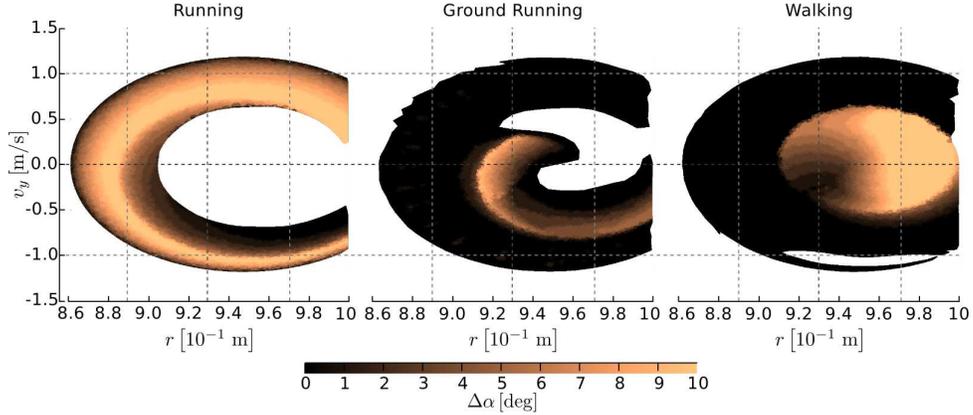}
\caption{\label{fig:DARegion} (Color online) Viability regions for each gait. The figures show the range of angles of attack that can be selected in each initial condition that allows the system give at least one more step. Colors indicate the size of the window, spanning from \unit{0}\degree to \unit{10}\degree.}
\end{figure*}

Fig. \ref{fig:DARegion} shows the $V^i(\Delta\alpha)$ regions for each gait $i$. Comparing with Fig.~\ref{fig:NMSRegion}, we see that in general long partial stability implies wider options for the angle of attack. Particularly, transitions are found near these regions of high viability and long partial stability, as will be described in the next section.

\begin{figure}[htb]

\includegraphics{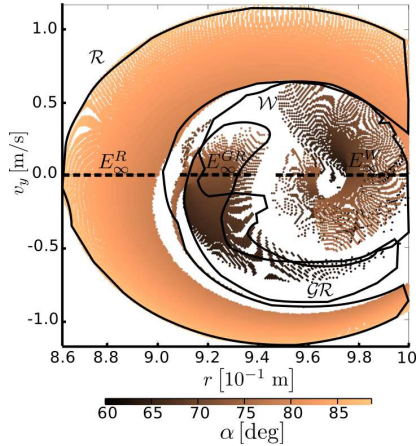}
\caption{\label{fig:OneStepRegion} (Color online) Points that can be mapped to stable regions in one step. The figures show the initial conditions that can be mapped to a small neighborhood of the stable region $E_\infty^i$,  $\vert v_y\vert < 1\times 10^{-3}$ ($v_y=0$, dashed horizontal lines). Color indicates the angle chosen. Regions $V^i(2\degree)$ are marked with solid lines.}
\end{figure}

Fig.~\ref{fig:OneStepRegion} shows one of the strongest results presented here. For each gait $i$, there is at least one angle of attack that maps the current state of the system into $E_\infty^i$, and this angle exists for an extense region of $\mathcal{S}$. This implies that if we consider control policies with variable angle of attack, almost any point in the section can be rendered stable. For this region the optimal control policy requires two angles: the first one maps the point to $E_\infty^i$; the second angle, keeps the system in this region. 

\subsection{Transition regions\label{sec:transitions}}

\begin{figure}[htb]
\includegraphics {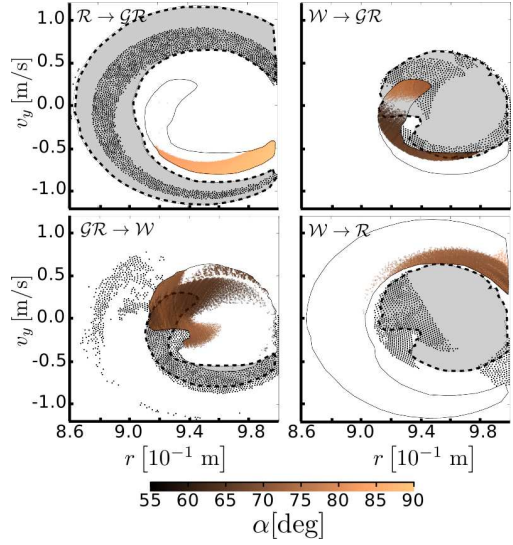}
\caption{\label{fig:TransRegion} (Color online) Transitions regions landing in $\Delta\alpha \geq 2\degree$. All the initial conditions that have a future inside the region with $\Delta\alpha \geq 2\degree$ of the objective gait are plotted with black dots. The same region of the starting gait is given as a reference and appears shaded. Colors in the objective region indicate the angle of attack used to perform the transition. Wherever two regions of different gaits intersect, the transition is automatically given.}
\end{figure}

As it was shown in the previous section, the only way of producing transitions between gaits is to put the system in a region with finite stability (due to the empty intersection of the $E_\infty^i$ regions reported in~\cite{JuergenRummel2009}, see Fig~\ref{fig:OneStepRegion}). In Fig.~\ref{fig:TransRegion} we show transitions starting at $E_n^i$ and arriving at $V^j\left(2\degree\right)$ for $i\neq j$ and $(i \rightarrow j)=\lbrace (R\rightarrow GR),(GR\rightarrow W),(W \rightarrow GR),(W \rightarrow R)\rbrace$. We show the transitions that will be used in the next example, however transitions between two any gaits are possible. It shall be noticed that wherever two regions of different gaits intersect, the transition is trivial.

\begin{figure}[htb]
\includegraphics{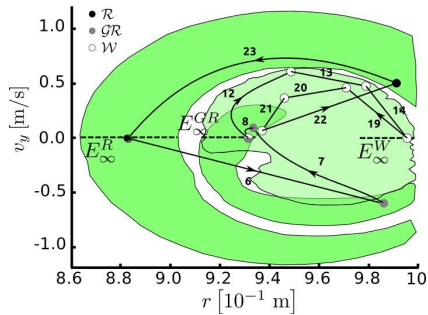}
\caption{\label{fig:Transition} (Color online) Transition sequence. The plot shows a trajectory with three transitions. The Regions $V^i\left(2\degree\right)$ are shown shaded with self-stable regions in dotted line. The arrows indicate the order of the sequence and the step number is given. The angle of attack sequence is given in (\ref{eq:angleseq}).}
\end{figure}

\begin{figure}[htb]
\includegraphics{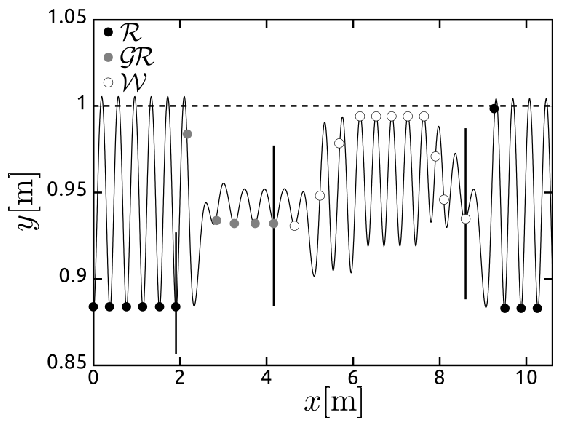}
\caption{\label{fig:TransitionTimeSeries} Transition time series. The figure shows the motion of the point mass\footnote{An animation of these transitions can be seen in http://www.ifi.uzh.ch/arvo/ailab/people/hamarti/GaitT.avi} in the plane is shown together with the crossing of the section (filled circles \ref{fig:Transition}). Transition points are indicated with a vertical line.}
\end{figure}

Finally, Fig.~\ref{fig:Transition} and Fig.~\ref{fig:TransitionTimeSeries} show one example of three transitions for a given initial condition. The trajectory has a total of $26$ steps and the angle sequence is 
\begin{equation}
\begin{split}
\alpha =\left( 81.886^5,88.500,62.400,72.350,71.100^3,71.000,\right. \\
 \left. 74.400,72.130,74.000^4,78.000^2,76.500,69.000,81.728^4 \right)&\end{split}
\label{eq:angleseq}
\end{equation}
\noindent where the exponent indicates how many times the angle was used. The path of the center of mass in the Cartesian plane is also shown in the figures.

All together we have shown that the SLIP model can be easily controlled to present transitions between gaits. To find transitions we must search for an intersection between the future of the starting region and the desired objective region. Depending how these regions are defined, it may be the case that multiple steps are required to achieve a successful transition.

\section{Discussion\label{sec:discussion}}
There are two important aspects regarding the viability regions. First, it is important to notice that $V^i(\Delta\alpha )$ enclose the $E_\infty^i$ region, and the points that can be mapped to stable regions in one step (Fig. \ref{fig:OneStepRegion}) .  Second, as it can be seen in Fig. \ref{fig:DARegion}, the bigger the range of the angle of attack is, the smaller the viability region is. We can take advantage of these properties to stabilize the system more easily. The selection of an appropriate $\Delta\alpha $ e.g. $2\degree $ defines a set of $V^i(\Delta\alpha) $ inside the section $\mathcal{S}$, where the controller has at least a range of $2\degree $ to select an appropriate angle of attack. Moreover, the agent can select conservative angles, step by step, to bring itself to the $E_\infty^i$ region (Fig. \ref{fig:TransRegion}).  

Despite the relief to the controller induced by the viability region, the selection of the $\Delta\alpha $ can generate regions that do not intersect; e.g. in Fig. \ref{fig:OneStepRegion} we can see that $V^i(2\degree) $ does not intersect any other region, which makes the gait transition more difficult to carry out. In order to cope with this situation, we look at the future of all the initial conditions in $E_n^i$. As it is presented in Fig. \ref{fig:TransRegion}, we found that there are some initial conditions, that under a set of angles of attack, are mapped from $E_n^i$ to $E_n^j$ (e.g.  $E_n^R$ to $E_n^{GR}$). What is also important is that the region where we can find these initial conditions are inside the viability region (Fig. \ref{fig:TransRegion}).

In these terms, the controller has two purposes. First, based on the state on the $\mathcal{S} $ section, it has to select the gait, and the angle of attack to keep the agent stable. Thus, the controller needs to have the knowledge of all the $V^R(\Delta\alpha) $, and the desired $\Delta\alpha $ to identify which gait has to be selected; the angle of attack can be selected based on the gait model. Second, the controller has to be able to produce gait transition when it is needed. Hence, the transition regions should be known by the controller and with a model of the gait, the angle of attack required can be selected. We expect that this approach can be used to handle uneven terrain, given that these irregularities can be modeled (under certain restrictions) as a change in energy.  

All these results are conditioned to the selection of the $\mathcal{S}$ section.  This means that we are analyzing the system in only one point in the whole trajectory.  From what we see in these results, in some regions the trajectories are very close. It would not be a surprise that these trajectories of $\gait{R}$, $\gait{W}$, and $\gait{GR}$ cross each other in another point along their continuous evolution, but given that we are looking just at the $\mathcal{S}$ section, this cannot be anticipated. Nevertheless, the selection of this section establishes the angle of attack as a natural control action to stabilize the system and to generate the transitions. 

\section{Conclusion\label{sec:conclusion}}

In the present study we have taken advantage of the perspective of hybrid dynamical systems to represent locomotion as a process generated by several charts. Although, this view makes evident a bigger set of connections among the charts, in this paper we take into account a small subset (s-chart to ff-chart, and s-chart to d-chart) which allow us to discover new alternatives to perform gait transitions. The development of the maps $\gait{W}_\alpha^1$, $\gait{GR}_\alpha^1$, $\gait{R}_\alpha^1$ is fundamental to identify important regions in the $\mathcal{S}$ section that bring the system to stable locomotion and to a gait transition.  The present results bring new ideas about plausible mechanisms that biped creatures could use to carry out gait transitions and stable locomotion.  These mechanisms exploit the passive dynamics of the system, which reduces the amount of energy needed to control the system.  These features are also present in biped machines with compliant legs, and as suggested in this paper, these mechanisms can be exploited to develop stable gaits and gait transitions. 

\section*{Acknowledgments}
Funding for this work has been supplied by SNSF project no. 122279 (From locomotion to cognition),  and by the European project no. ICT-2007.2.2 (ECCEROBOT). Additionally, the research leading to these results has received funding from the European Community's Seventh Framework Programme FP7/2007-2013-Challenge 2-Cognitive Systems, Interaction, Robotics- under grant agreement No 248311-AMARSi.

\bibliography{references}

\begin{thebibliography}{10}%
\makeatletter
\providecommand \@ifxundefined [1]{%
 \ifx #1\undefined \expandafter \@firstoftwo
 \else \expandafter \@secondoftwo
\fi
}%
\providecommand \@ifnum [1]{%
 \ifnum #1\expandafter \@firstoftwo
 \else \expandafter \@secondoftwo
\fi
}%
\providecommand \enquote [1]{``#1''}%
\providecommand \bibnamefont  [1]{#1}%
\providecommand \bibfnamefont [1]{#1}%
\providecommand \citenamefont [1]{#1}%
\providecommand\href[0]{\@sanitize\@href}%
\providecommand\@href[1]{\endgroup\@@startlink{#1}\endgroup\@@href}%
\providecommand\@@href[1]{#1\@@endlink}%
\providecommand \@sanitize [0]{\begingroup\catcode`\&12\catcode`\#12\relax}%
\@ifxundefined \pdfoutput {\@firstoftwo}{%
 \@ifnum{\z@=\pdfoutput}{\@firstoftwo}{\@secondoftwo}%
}{%
 \providecommand\@@startlink[1]{\leavevmode\special{html:<a href="#1">}}%
 \providecommand\@@endlink[0]{\special{html:</a>}}%
}{%
 \providecommand\@@startlink[1]{%
  \leavevmode
  \pdfstartlink
   attr{/Border[0 0 1 ]/H/I/C[0 1 1]}%
   user{/Subtype/Link/A<</Type/Action/S/URI/URI(#1)>>}%
  \relax
 }%
 \providecommand\@@endlink[0]{\pdfendlink}%
}%
\providecommand \url  [0]{\begingroup\@sanitize \@url }%
\providecommand \@url [1]{\endgroup\@href {#1}{\urlprefix}}%
\providecommand \urlprefix [0]{URL }%
\providecommand \Eprint[0]{\href }%
\@ifxundefined \urlstyle {%
  \providecommand \doi [1]{doi:\discretionary{}{}{}#1}%
}{%
  \providecommand \doi [0]{doi:\discretionary{}{}{}\begingroup
  \urlstyle{rm}\Url }%
}%
\providecommand \doibase [0]{http://dx.doi.org/}%
\providecommand \Doi[1]{\href{\doibase#1}}%
\providecommand \bibAnnote [3]{%
  \BibitemShut{#1}%
  \begin{quotation}\noindent
    \textsc{Key:}\ #2\\\textsc{Annotation:}\ #3%
  \end{quotation}%
}%
\providecommand \bibAnnoteFile [2]{%
  \IfFileExists{#2}{\bibAnnote {#1} {#2} {\input{#2}}}{}%
}%
\providecommand \typeout [0]{\immediate \write \m@ne }%
\providecommand \selectlanguage [0]{\@gobble}%
\providecommand \bibinfo [0]{\@secondoftwo}%
\providecommand \bibfield [0]{\@secondoftwo}%
\providecommand \translation [1]{[#1]}%
\providecommand \BibitemOpen[0]{}%
\providecommand \bibitemStop [0]{}%
\providecommand \bibitemNoStop [0]{.\EOS\space}%
\providecommand \EOS [0]{\spacefactor3000\relax}%
\providecommand \BibitemShut [1]{\csname bibitem#1\endcsname}%
\bibitem{HolmesSIAM.06}%
  \BibitemOpen
  \bibfield{author}{%
  \bibinfo {author} {\bibfnamefont{P.}~\bibnamefont{Holmes}}, \bibinfo {author}
  {\bibfnamefont{R.~J.}\ \bibnamefont{Full}}, \bibinfo {author}
  {\bibfnamefont{D.}~\bibnamefont{Koditschek}},\ and\ \bibinfo {author}
  {\bibfnamefont{J.}~\bibnamefont{Guckenheimer}},\ }%
  \bibfield{journal}{%
  \bibinfo {journal} {SIAM Rev.}\ }%
  \textbf{\bibinfo {volume} {48}},\ \bibinfo {pages} {207} (\bibinfo {year}
  {2006}),\ ISSN \bibinfo {issn} {0036-1445}%
  \bibAnnoteFile{NoStop}{HolmesSIAM.06}%
\bibitem{Mochon1980}%
  \BibitemOpen
  \bibfield{author}{%
  \bibinfo {author} {\bibfnamefont{S.}~\bibnamefont{Mochon}}\ and\ \bibinfo
  {author} {\bibfnamefont{T.~A.}\ \bibnamefont{McMahon}},\ }%
  \bibfield{journal}{%
  \Doi{10.1016/0021-9290(80)90007-X}{\bibinfo {journal} {J. Biomech.}}\ }%
  \textbf{\bibinfo {volume} {13}},\ \bibinfo {pages} {49} (\bibinfo {year}
  {1980}),\ ISSN \bibinfo {issn} {00219290}%
  \bibAnnoteFile{NoStop}{Mochon1980}%
\bibitem{Geyer2006}%
  \BibitemOpen
  \bibfield{author}{%
  \bibinfo {author} {\bibfnamefont{H.}~\bibnamefont{Geyer}}, \bibinfo {author}
  {\bibfnamefont{A.}~\bibnamefont{Seyfarth}},\ and\ \bibinfo {author}
  {\bibfnamefont{R.}~\bibnamefont{Blickhan}},\ }%
  \bibfield{journal}{%
  \Doi{10.1098/rspb.2006.3637}{\bibinfo {journal} {P. Roy. Soc. B - Biol.
  Sci.}}\ }%
  \textbf{\bibinfo {volume} {273}},\ \bibinfo {pages} {2861} (\bibinfo {month}
  {Nov.}\ \bibinfo {year} {2006}),\ ISSN \bibinfo {issn} {0962-8452}%
  \bibAnnoteFile{NoStop}{Geyer2006}%
\bibitem{Guckenheimer95}%
  \BibitemOpen
  \bibfield{author}{%
  \bibinfo {author} {\bibfnamefont{J.}~\bibnamefont{Guckenheimer}}\ and\
  \bibinfo {author} {\bibfnamefont{S.}~\bibnamefont{Johnson}},\ }%
  in\ \emph{\bibinfo {booktitle} {Hybrid Systems II}}\ (\bibinfo {publisher}
  {Springer-Verlag},\ \bibinfo {address} {London, UK},\ \bibinfo {year}
  {1995})\ pp.\ \bibinfo {pages} {202--225},\ ISBN \bibinfo {isbn}
  {3-540-60472-3}%
  \bibAnnoteFile{NoStop}{Guckenheimer95}%
\bibitem{Cortes2008}%
  \BibitemOpen
  \bibfield{author}{%
  \bibinfo {author} {\bibfnamefont{J.}~\bibnamefont{Cortes}},\ }%
  \bibfield{journal}{%
  \Doi{10.1109/MCS.2008.919306}{\bibinfo {journal} {IEEE Contr. Sys. Mag.}}\ }%
  \textbf{\bibinfo {volume} {28}},\ \bibinfo {pages} {36} (\bibinfo {month}
  {Jun.}\ \bibinfo {year} {2008}),\ ISSN \bibinfo {issn} {0272-1708}%
  \bibAnnoteFile{NoStop}{Cortes2008}%
\bibitem{McGeer1990}%
  \BibitemOpen
  \bibfield{author}{%
  \bibinfo {author} {\bibfnamefont{T.}~\bibnamefont{McGeer}},\ }%
  \bibfield{journal}{%
  \Doi{10.1177/027836499000900206}{\bibinfo {journal} {Int. J. Robot. Res.}}\
  }%
  \textbf{\bibinfo {volume} {9}},\ \bibinfo {pages} {62} (\bibinfo {month}
  {Apr.}\ \bibinfo {year} {1990}),\ ISSN \bibinfo {issn} {0278-3649}%
  \bibAnnoteFile{NoStop}{McGeer1990}%
\bibitem{Collins2001}%
  \BibitemOpen
  \bibfield{author}{%
  \bibinfo {author} {\bibfnamefont{S.~H.}\ \bibnamefont{Collins}},\ }%
  \bibfield{journal}{%
  \Doi{10.1177/02783640122067561}{\bibinfo {journal} {Int. J. Robot. Res.}}\ }%
  \textbf{\bibinfo {volume} {20}},\ \bibinfo {pages} {607} (\bibinfo {month}
  {Jul.}\ \bibinfo {year} {2001}),\ ISSN \bibinfo {issn} {0278-3649}%
  \bibAnnoteFile{NoStop}{Collins2001}%
\bibitem{MartijnWisseandJanvanFrankenhuyzen2006}%
  \BibitemOpen
  \bibfield{author}{%
  \bibinfo {author} {\bibfnamefont{M.}~\bibnamefont{Wisse}}\ and\ \bibinfo
  {author} {\bibfnamefont{J.~V.}\ \bibnamefont{Frankenhuyzen}},\ }%
  in\ \emph{\bibinfo {booktitle} {Adaptive Motion of Animals and Machines}}\
  (\bibinfo {publisher} {Springer-Verlag},\ \bibinfo {address} {Tokyo},\
  \bibinfo {year} {2006})\ pp.\ \bibinfo {pages} {143--154},\ ISBN \bibinfo
  {isbn} {4-431-24164-7}%
  \bibAnnoteFile{NoStop}{MartijnWisseandJanvanFrankenhuyzen2006}%
\bibitem{Collins2005}%
  \BibitemOpen
  \bibfield{author}{%
  \bibinfo {author} {\bibfnamefont{S.~H.}\ \bibnamefont{Collins}}, \bibinfo
  {author} {\bibfnamefont{A.}~\bibnamefont{Ruina}}, \bibinfo {author}
  {\bibfnamefont{R.}~\bibnamefont{Tedrake}},\ and\ \bibinfo {author}
  {\bibfnamefont{M.}~\bibnamefont{Wisse}},\ }%
  \bibfield{journal}{%
  \Doi{10.1126/science.1107799}{\bibinfo {journal} {Science}}\ }%
  \textbf{\bibinfo {volume} {307}},\ \bibinfo {pages} {1082} (\bibinfo {month}
  {Feb.}\ \bibinfo {year} {2005}),\ ISSN \bibinfo {issn} {1095-9203}%
  \bibAnnoteFile{NoStop}{Collins2005}%
\bibitem{Geng2006}%
  \BibitemOpen
  \bibfield{author}{%
  \bibinfo {author} {\bibfnamefont{T.}~\bibnamefont{Geng}}, \bibinfo {author}
  {\bibfnamefont{B.}~\bibnamefont{Porr}},\ and\ \bibinfo {author}
  {\bibfnamefont{F.}~\bibnamefont{Worgotter}},\ }%
  \bibfield{journal}{%
  \Doi{10.1162/neco.2006.18.5.1156}{\bibinfo {journal} {Neural Comput.}}\ }%
  \textbf{\bibinfo {volume} {18}},\ \bibinfo {pages} {1156} (\bibinfo {month}
  {May}\ \bibinfo {year} {2006}),\ ISSN \bibinfo {issn} {0899-7667}%
  \bibAnnoteFile{NoStop}{Geng2006}%
\bibitem{Rummel2010}%
  \BibitemOpen
  \bibfield{author}{%
  \bibinfo {author} {\bibfnamefont{J.}~\bibnamefont{Rummel}}, \bibinfo {author}
  {\bibfnamefont{Y.}~\bibnamefont{Blum}}, \bibinfo {author}
  {\bibfnamefont{H.~M.}\ \bibnamefont{Maus}}, \bibinfo {author}
  {\bibfnamefont{C.}~\bibnamefont{Rode}},\ and\ \bibinfo {author}
  {\bibfnamefont{A.}~\bibnamefont{Seyfarth}},\ }%
  in\ \emph{\bibinfo {booktitle} {IEEE Int. Conf. Robot. (ICRA)}}\ (\bibinfo
  {publisher} {IEEE},\ \bibinfo {year} {2010})\ pp.\ \bibinfo {pages}
  {5250--5255},\ ISBN \bibinfo {isbn} {978-1-4244-5038-1}%
  \bibAnnoteFile{NoStop}{Rummel2010}%
\bibitem{JuergenRummel2009}%
  \BibitemOpen
  \bibfield{author}{%
  \bibinfo {author} {\bibfnamefont{J.}~\bibnamefont{Rummel}}, \bibinfo {author}
  {\bibfnamefont{Y.}~\bibnamefont{Blum}},\ and\ \bibinfo {author}
  {\bibfnamefont{A.}~\bibnamefont{Seyfarth}},\ }%
  in\ \emph{\bibinfo {booktitle} {Autonome Mobile Systeme}}\ (\bibinfo
  {publisher} {Springer},\ \bibinfo {address} {Berlin, Heidelberg},\ \bibinfo
  {year} {2009})\ pp.\ \bibinfo {pages} {89--96},\ ISBN \bibinfo {isbn}
  {978-3-642-10283-7}%
  \bibAnnoteFile{NoStop}{JuergenRummel2009}%
\bibitem{Piiroinen08}%
  \BibitemOpen
  \bibfield{author}{%
  \bibinfo {author} {\bibfnamefont{P.~T.}\ \bibnamefont{Piiroinen}}\ and\
  \bibinfo {author} {\bibfnamefont{Y.~A.}\ \bibnamefont{Kuznetsov}},\ }%
  \bibfield{journal}{%
  \Doi{10.1145/1356052.1356054}{\bibinfo {journal} {ACM T. Math. Software}}\ }%
  \textbf{\bibinfo {volume} {34}},\ \bibinfo {pages} {1} (\bibinfo {month}
  {May}\ \bibinfo {year} {2008}),\ ISSN \bibinfo {issn} {00983500}%
  \bibAnnoteFile{NoStop}{Piiroinen08}%
\bibitem{octave2002}%
  \BibitemOpen
  \bibfield{author}{%
  \bibinfo {author} {\bibfnamefont{J.~W.}\ \bibnamefont{Eaton}},\ }%
  \emph{\bibinfo {title} {GNU Octave Manual}}\ (\bibinfo {publisher} {Network
  Theory Limited, http://www.octave.org},\ \bibinfo {year} {2002})\ ISBN
  \bibinfo {isbn} {0-9541617-2-6},\ \url{http://www.octave.org}%
  \bibAnnoteFile{NoStop}{octave2002}%
\bibitem{matplotlib}%
  \BibitemOpen
  \bibfield{author}{%
  \bibinfo {author} {\bibfnamefont{J.~D.}\ \bibnamefont{Hunter}},\ }%
  \bibfield{journal}{%
  \Doi{http://doi.ieeecomputersociety.org/10.1109/MCSE.2007.55}{\bibinfo
  {journal} {Computing in Science and Engineering}}\ }%
  \textbf{\bibinfo {volume} {9}},\ \bibinfo {pages} {90} (\bibinfo {year}
  {2007}),\ ISSN \bibinfo {issn} {1521-9615}%
  \bibAnnoteFile{NoStop}{matplotlib}%
\end{thebibliography}%

\end{document}